\documentclass[10pt,twocolumn,letterpaper]{article}

\usepackage{cvpr}
\usepackage{times}
\usepackage{epsfig}
\usepackage{graphicx}
\usepackage{amsmath}
\usepackage{amssymb}

\usepackage[breaklinks=true,bookmarks=false]{hyperref}

\usepackage{amsthm}

\newcommand{\argmax}{\mathop{\rm argmax}\limits}

\newcommand{\figref}[1]{Fig.~\ref{#1}}
\newcommand{\tabref}[1]{Table.~\ref{#1}}

\newcommand{\secref}[1]{Sec.~\ref{#1}}

\renewcommand{\thefootnote}{\alph{footnote}}

\newcommand{\astfootnote}[1]{
	\let\oldthefootnote=\thefootnote
	\setcounter{footnote}{0}
	\renewcommand{\thefootnote}{\fnsymbol{footnote}}
	\footnote{#1}
	\let\thefootnote=\oldthefootnote
}

\cvprfinalcopy 

\def\cvprPaperID{****} 

\begin{document}
	
	\title{Key Instance Selection for Unsupervised Video Object Segmentation}
	
	\author{Donghyeon Cho\footnotemark\\
		SK T-Brain\\
		\and
		Sungeun Hong\footnotemark[\value{footnote}]\\
		SK T-Brain\\
		\and
		Sungil Kang\footnotemark[\value{footnote}]\\
		SK T-Brain\\
		\and
		Jiwon Kim\\
		SK T-Brain\\
	}
	
	\maketitle
	
	\begin{abstract}
		This paper proposes key instance selection based on video saliency covering objectness and dynamics for unsupervised video object segmentation (UVOS).
		Our method takes frames sequentially and extracts object proposals with corresponding masks for each frame.  
		We link objects according to their similarity until the $M$-th frame and then assign them unique IDs (i.e., instances).
		Similarity measure takes into account multiple properties such as ReID descriptor, expected trajectory, and semantic co-segmentation result.
		After $M$-th frame, we select $K$ IDs based on video saliency and frequency of appearance; then only these key IDs are tracked through the remaining frames.
		Thanks to these technical contributions, our results are ranked third on the leaderboard of UVOS DAVIS challenge.
		
	\end{abstract}
	\footnotetext{*They are equal contributors to this work.}
	\section{Introduction}
	Given a mask in the first frame, semi-supervised video object segmentation (SVOS) is a task of generating masks in the subsequent frames. After the first SVOS challenge~\cite{Perazzi_cvpr2016}, the SVOS has been gradually getting attention. Also, datasets are continuously updated~\cite{DAVIS_2017} or newly constructed~\cite{Xu_youtubeVOS}. Recently, interactive video object segmentation (IVOS)~\cite{DAVIS_2018} and unsupervised video object segmentation (UVOS)~\cite{DAVIS_2019} were introduced as new challenges. This paper tackles UVOS which does not require any human supervision.
	
\begin{figure}
		\begin{center}
			\def\arraystretch{0.5}
			\begin{tabular}{@{}c@{\hskip 0.01\linewidth}c@{\hskip 0.01\linewidth}c@{\hskip 0.01\linewidth}c@{\hskip 0.01\linewidth}c}
				
				\includegraphics[width=0.32\linewidth]{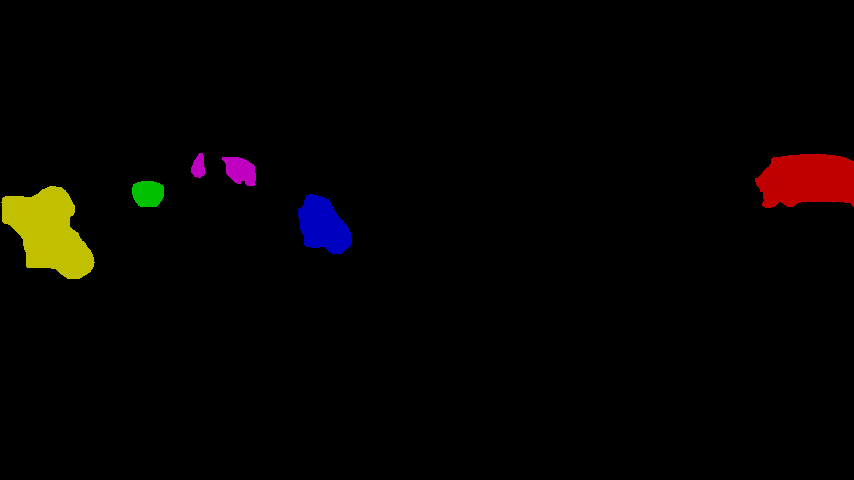} &
				\includegraphics[width=0.32\linewidth]{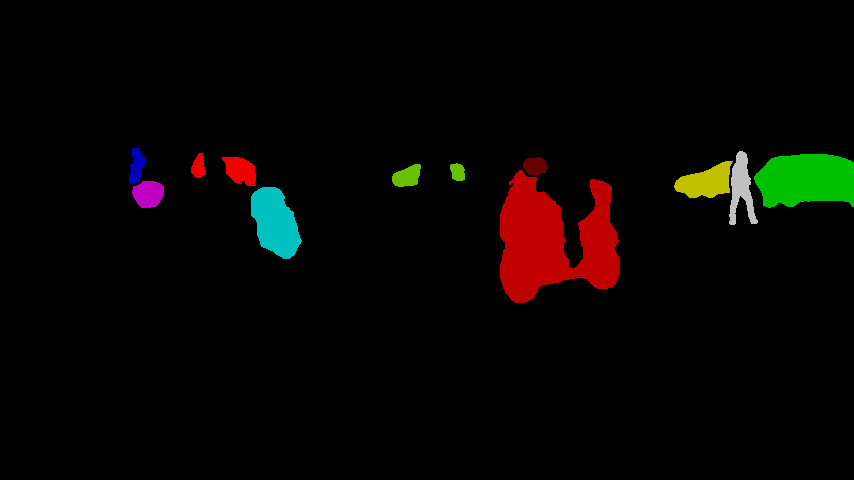} &
				\includegraphics[width=0.32\linewidth]{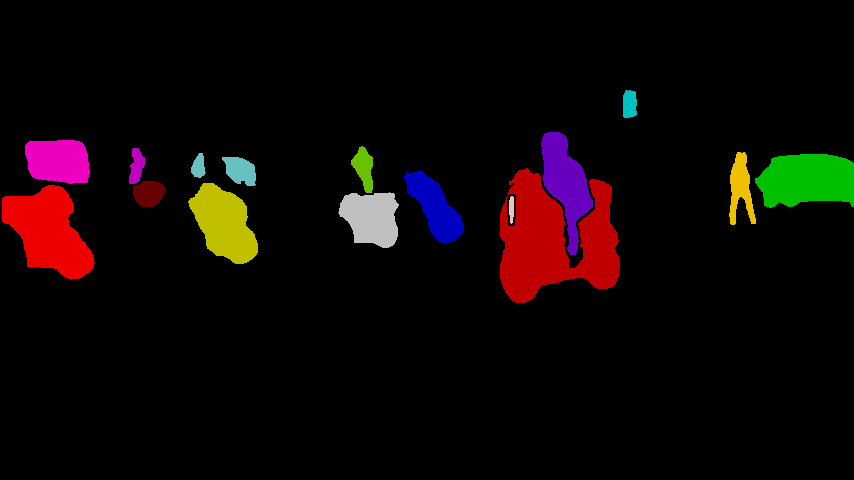}  \\
				\includegraphics[width=0.32\linewidth]{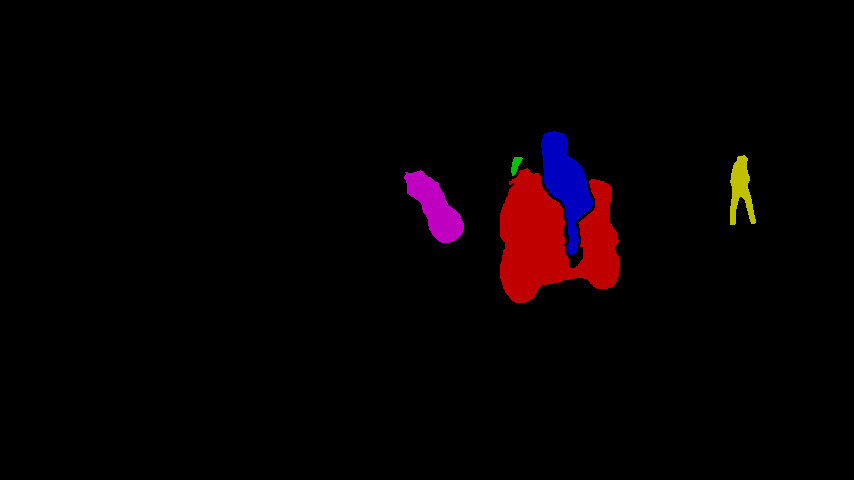} &
				\includegraphics[width=0.32\linewidth]{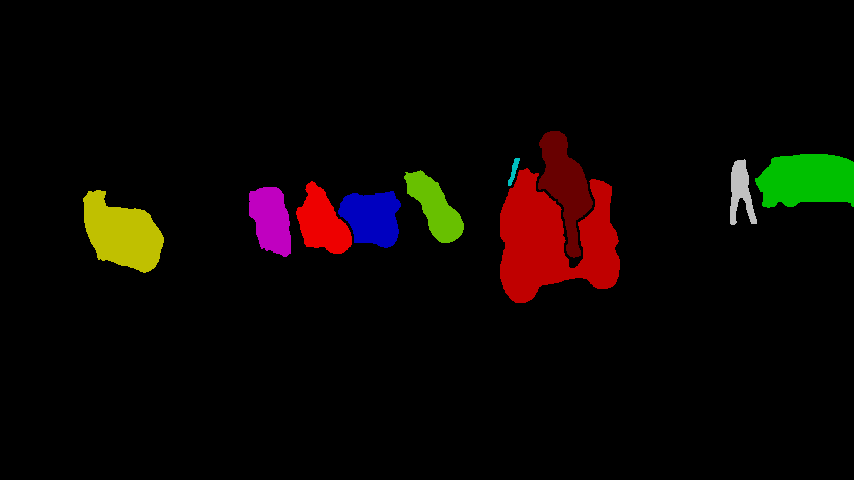} &
				\includegraphics[width=0.32\linewidth]{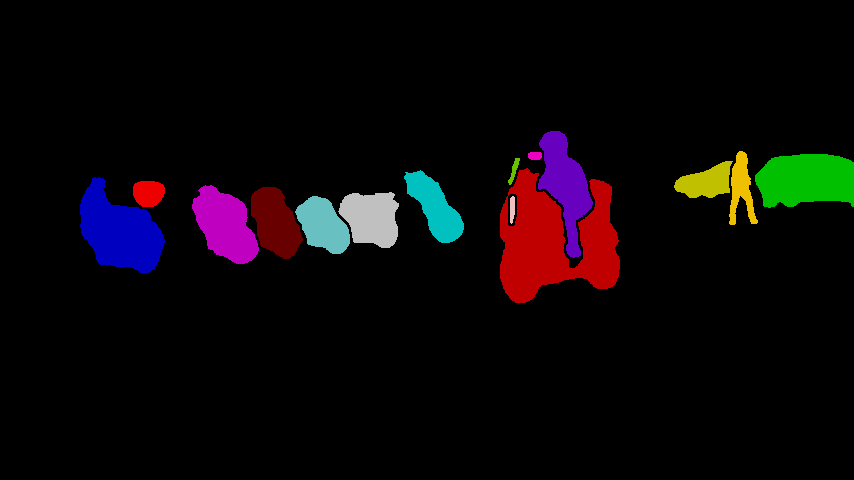}  \\
				{(a) $K$=5 } & {(b) $K$=10 } & {(c) $K$=15} 
			\end{tabular}
		\end{center}
		\caption{Segmentation results with respect to $K$ instances on a `scooter-black' video. Top: random instance selection. \: Bottom: our key instance selection.}
		\label{fig:topk}
	\end{figure}

	The basic approach of the UVOS is to estimate a mask of the first frame, then use it to apply conventional SVOS methods~\cite{Luiten_cvprw2018,Luiten_2018accv}. However, not only is it greatly influenced by the results of the first frame, it is not guaranteed that there are all targets in the first frame.  To resolve these problems, we can  continuously assign a new ID (i.e., instance) to an object that satisfies certain criteria. However, this increases not only the number of IDs indiscriminately but also time complexity.  Therefore, it is recommended to fix the appropriate $K$, the maximum number of IDs. In this paper, we propose a method to select $K$ IDs by exploring video saliency of several frames at the beginning of the video. Compared to random $K$ instance selection, our method can capture main objects well, even in case of small $K$ as shown in~\figref{fig:topk}-(a). 
	
	
	The main contributions of this paper are as follows. Compared to adding new IDs continuously until the end of the video, our key instance selection approach significantly reduces time as well as memory complexity.
	Also, our model can capture regions of importance better than random instance selection.
	When we measure the similarity between assigned
	IDs and extracted proposals, we use all set of positive ReID descriptors for each ID.
	Unlike conventional methods that use optical flow~\cite{Ilg_cvpr2017} to handle large appearance changes in a frame sequence, we use semantic co-segmentation~\cite{oh_2018cvpr,Oh_cvprw2018}. 
	Finally, we automatically search  hyperparameters through Meta AI system developed by T-Brain under scalable environments (e.g., 144 GPUs).

	\section{Method}
	\label{sec:pipeline}
\begin{figure}
		\begin{center}
			\includegraphics[width=0.99\linewidth]{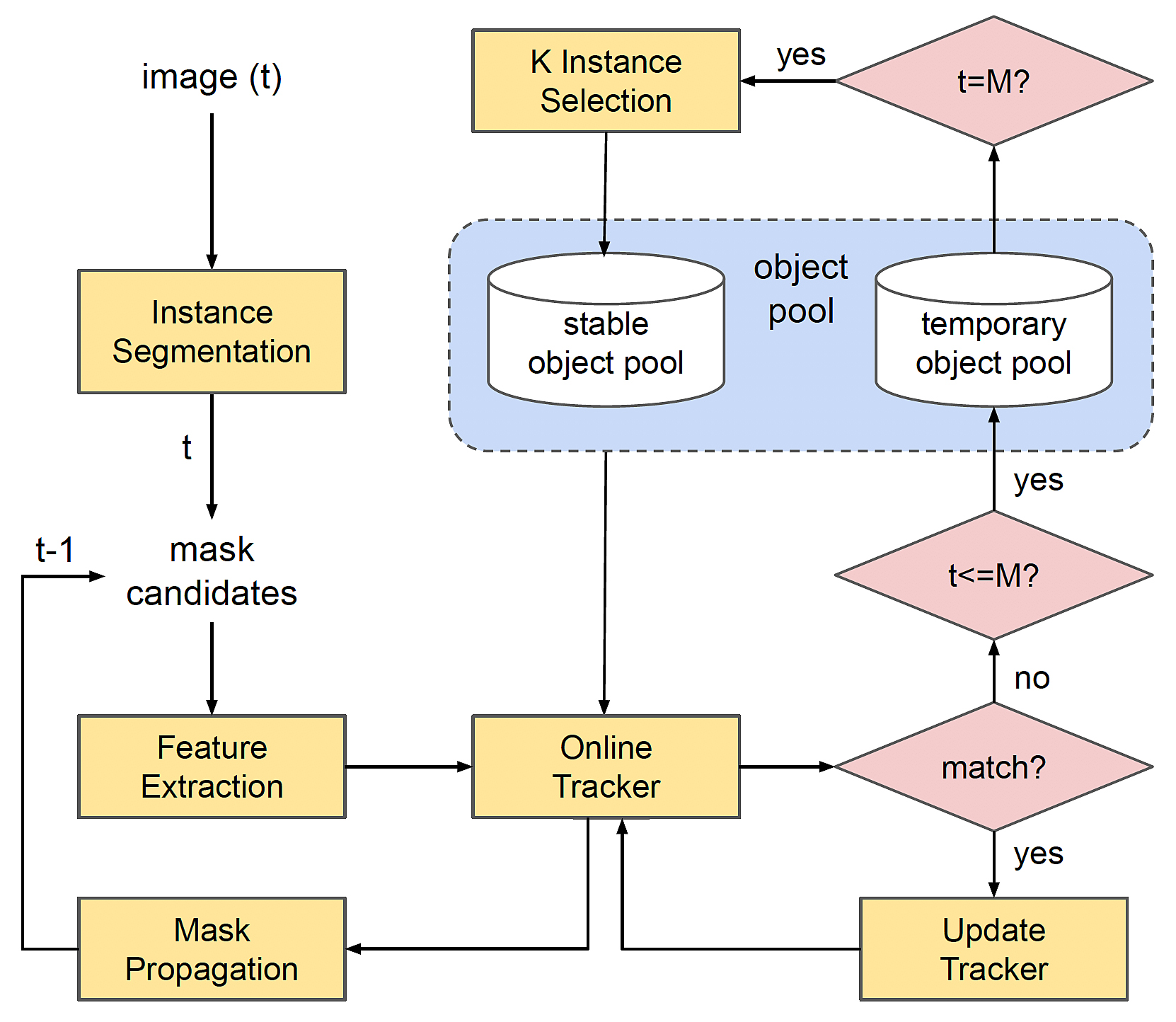} \\
		\end{center}
		\caption{Overview of the proposed method.}
		\label{fig:overview}
	\end{figure}

	Our objective is to assign IDs to the proposals without additional inputs such as a mask of the first frame. 
	For this, we use an object pool that manages assigning, adding, and deleting IDs.
	\figref{fig:overview} illustrates the overview of the proposed method.
	
	As a first step, we perform instance segmentation on the current frame and propagate the masks obtained in the previous frame.
	We then extract features and assign IDs to candidates that satisfy the criteria of the online tracker linked with object pool.
	Here, if the candidate matches the existing ID in the object pool, we assign the matched ID to the candidate, and the online tracker is updated; otherwise, we assign a new ID to the candidate, then this new ID is added to the object pool.
	Adding a new ID is performed only up to the $M$-th frame.
	At the $M$-th frame, $K$ instances among the accumulated IDs are selected and finally, those selected IDs are tracked in the remaining frames.

	\paragraph{Candidate Generation:}
	Given a frame at time  $I^{t}$ ($t \in T$), we perform instance segmentation to get bounding box  and mask by using  Mask R-CNN~\cite{He_iccv2017} and DeepLab~\cite{Chen_eccv2018}.
	Motion blur or occlusion, which often occurs in videos, can result in poor segmentation results for certain frames.
	To address this issue, we use masks propagated from the previous frame as mask candidates. 
	Concretely, we utilize RGMP~\cite{oh_2018cvpr,Oh_cvprw2018} in our experiments.

	\paragraph{ID Assignment:}
	To assign an ID to each candidate, we compute specific scores by comparing the candidates and the registered IDs in the object pool.
	The first score is $S_{iou}(l,n)$ which is IOU between a mask from ID ($l \in L$) and a mask of the candidate ($n \in N$).
	Here, $L$ refers to the number of IDs registered in the object pool and  $N$ is the number of candidates.
	We omit notation time $t$ for simplicity. 
	The second score is $S_{traj}(l,n)$ that measures how far the candidate is from the predicted bounding box of ID as
	\begin{eqnarray}
	S_{traj}(l,n) =1 - \min{(\frac{dot(\vec{b}_{l},\vec{b}_{n})}{\alpha_{traj}}, 1)},
	\label{eq:track}
	\end{eqnarray}
	where $\alpha_{traj}$ is a normalization factor. Also, $\vec{b}_{l}$ and $\vec{b}_{n}$  are vectors from bounding boxes of ID and candidate, respectively. The third score $S_{reid}(l,n)$ considers a distance of ReID descriptors~\cite{Osep_2017} between ID $l$ and candidate $n$. Unlike~\cite{Luiten_cvprw2018,Luiten_2018accv}, our method use the nearest ReID descriptor among the all set of positive ReID descriptors of ID $l$ as
	\begin{eqnarray}
	S_{reid}(l,n) = 1 - \min{(\frac{\min_{j} ||d_{l}(j) - r_{n}||}{\alpha_{reid}}, 1)},
	\label{eq:reid}
	\end{eqnarray}
	where $r_{n}$ is a ReID descriptor for candidate $n$ and $d_{l}(\cdot)$ are positive ReID descriptors for  ID $l$. The last score is a relative ReID score $S_{rel}(l,n)$:
	\begin{eqnarray}
	S_{rel}(l,n) = \frac{S_{reid}(l,n)}{\max_{l}S_{reid}(l,n)}.
	\label{eq:relative}
	\end{eqnarray}
	Total score is weighted summation of above four scores:
	\begin{eqnarray}
	\begin{split}
	S_{total}(l,n) = w_{iou} \cdot S_{iou}(l,n)  + w_{traj} \cdot S_{traj}(l,n) &\\ + w_{reid} \cdot S_{reid}(l,n)  + w_{rel} \cdot S_{rel}(l,n),
	\end{split}
	\label{eq:total}
	\end{eqnarray}
	where $w_{iou}$, $w_{traj}$, $w_{reid}$, and $w_{rel}$ are weight factors of each term. Finally, we assign ID $l$ to the candidate object as follows:
	\begin{eqnarray}
	\hat{n}= 
	\begin{cases}
	\argmax_{n} S_{total}(l,n),& \text{if } \geq \tau_{c}\\
	\mathrm{None},              & \text{otherwise}
	\end{cases}
	,
	\label{eq:assign}
	\end{eqnarray}
	where $\tau_{c=\left \{ 1,2 \right \}}$ is a threshold value for cutting off object with low confidence. Before selecting the $K$ instances ($t<=M$) as in~\secref{sec:K_selection}, $c$ is 1, and after that $c$ is 2. If one of candidates is assigned to ID $l$, then $d_{\hat{n}}$ is added to pool of positive ReID descriptors for ID $l$. Also $\vec{b}_{l}$ is updated by using $\vec{b}_{\hat{n}}$.
	
	\begin{figure}
		\begin{center}
			\def\arraystretch{0.5}
			\begin{tabular}{@{}c@{\hskip 0.01\linewidth}c@{\hskip 0.01\linewidth}c@{\hskip 0.01\linewidth}c@{\hskip 0.01\linewidth}c}
				
				\includegraphics[width=0.32\linewidth]{figures/sal/img1.jpg} &
				\includegraphics[width=0.32\linewidth]{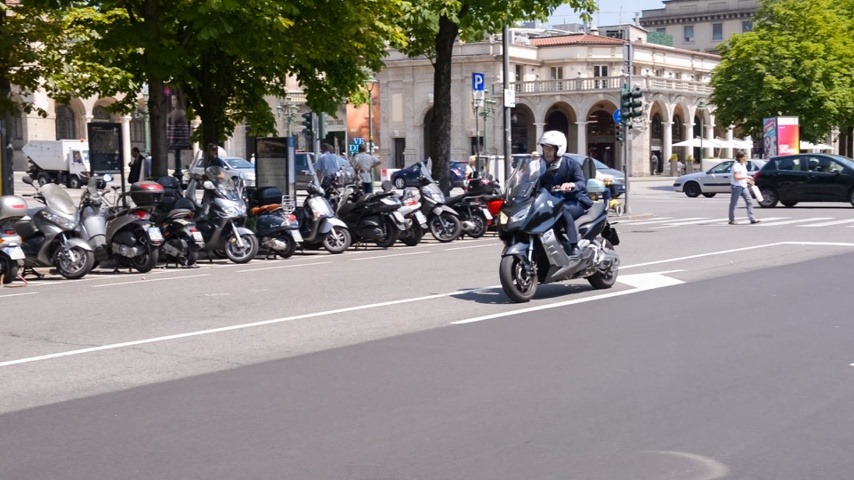} &
				\includegraphics[width=0.32\linewidth]{figures/sal/img3.jpg}  \\
				\includegraphics[width=0.32\linewidth]{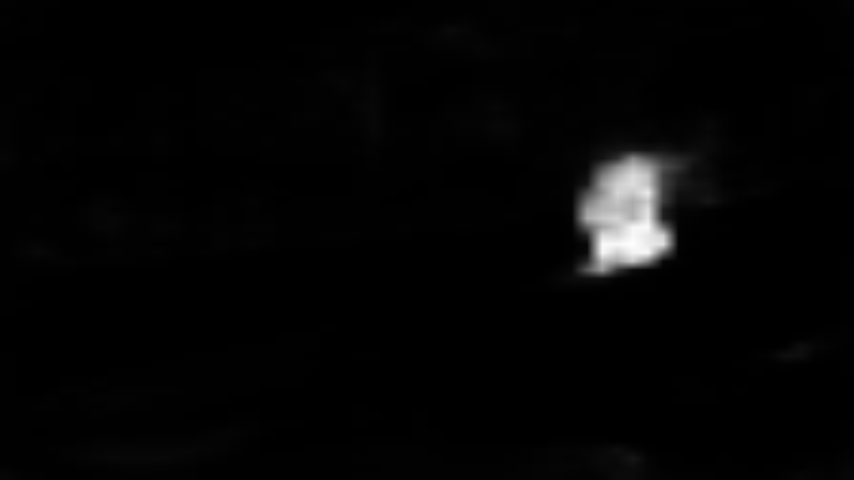} &
				\includegraphics[width=0.32\linewidth]{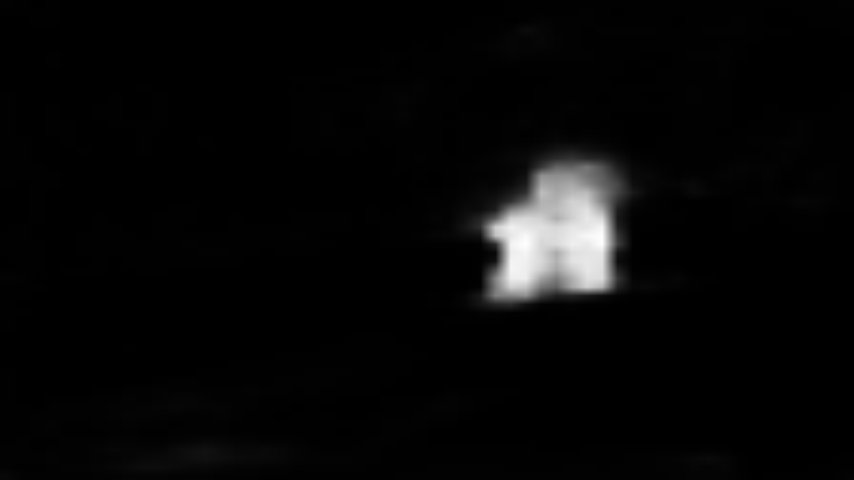} &
				\includegraphics[width=0.32\linewidth]{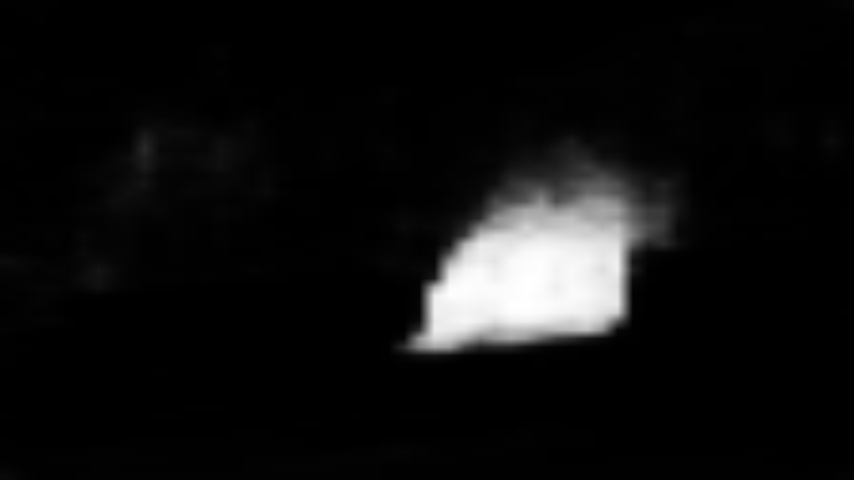}
			\end{tabular}
		\end{center}
		\caption{Video saliency results. Note that  a motorcycle with motion is only focused among a lot of objects in a scene.}
		\label{fig:video_sal}
	\end{figure}

	\begin{table*}[t]
		\centering
		\resizebox{1\linewidth}{!}
		{%
			\begin{tabular}{  l | cc cc cc cc cc cc cc cc}
				\hline
				Measure  && RWTH Vision && Oxford-CASIA && \textbf{SK T-Brain (ours)} && UVOS-test && RWTH Vision 2 && ZX\_VIP && VIG && UOC-UPC-BSC \\
				\hline
				\hline
				Ranking     && 1  && 2 && 3 && 4  && 5 && 6 && 7  && 8\\
				Global Mean  $\uparrow$   && 0.564  && 0.562 && 0.516 && 0.504  && 0.481 && 0.471 && 0.448 && 0.412\\
				\hline
				\hline
				$\mathcal{J}$ Mean    $\uparrow$   && 0.564  && 0.535 && 0.487  && 0.475  && 0.460 && 0.435 && 0.422  && 0.379\\
				$\mathcal{J}$ Recall   $\uparrow$  && 0.609  && 0.613 && 0.551 && 0.542   && 0.514  && 0.490 && 0.476 && 0.413\\
				$\mathcal{J}$ Decay  $\downarrow$ && 0.015  && -0.021 && 0.040 && 0.032  && 0.084 && 0.035 && 0.035 && 0.076\\
				\hline 
				\hline
				$\mathcal{F}$ Mean    $\uparrow$   && 0.594  && 0.590 && 0.545 && 0.533  && 0.503 && 0.506 && 0.474  && 0.444\\
				$\mathcal{F}$ Recall   $\uparrow$  && 0.641  && 0.632  && 0.594 && 0.569  && 0.538  && 0.543 && 0.506 && 0.473 \\
				$\mathcal{F}$ Decay $\downarrow$  && 0.058  && 0.001 && 0.077  && 0.055   && 0.118 && 0.067 && 0.068 && 0.117\\
				\hline 
			\end{tabular}
		}
		\linebreak
		\caption{Segmentation results on UVOS DAVIS challenge dataset.  }
		\label{tab:results_challenge}
	\end{table*}
	
	\paragraph{Key Instances Selection:}
	\label{sec:K_selection}
	Basically, the pipeline mentioned above is iterated at each frame. There is ID addition process at the beginning of the video. Especially, in the first frame, object candidates with high confidence are added as new IDs. From the second frame and $M$-th frame, new IDs are added when object candidates have high objectness score and are not overlapped with objects assigned to existing IDs. After $M$-th frame, we select at most $K$ IDs. As selection criteria, we use weighted summation of video saliency score~\cite{Wang_tip2018} and frequency of each ID as
	\begin{eqnarray}
	S_{sel}(l) = w_{sal} \cdot S_{sal}(l)  + w_{freq} \cdot S_{freq}(l),
	\label{eq:selection}
	\end{eqnarray}
	where $w_{sal}$ and $w_{freq}$ are weight factors of each term. $S_{sal}(l)$ and $S_{freq}(l)$ are computed over frames. \figref{fig:video_sal} shows effectiveness of video saliency based approach. The frequency of each ID means the number of each ID's appearance up to the $M$-th frame. Because there are several IDs that are not connected to any proposal at a certain time by $\tau_{1}$, $S_{freq}(l)$ can vary by  ID.

	\paragraph{Hyperparameter Search:}
	We perform hyperparameter search for $w_{iou}$, $w_{traj}$, $w_{reid}$, $w_{rel}$, $\tau_{1}$, $\tau_{2}$, $w_{sal}$ and $w_{freq}$ on DAVIS validation dataset. Global mean ($\mathcal{J}$$\&$$\mathcal{F}$) of results is $0.599$ and searched weighting factors are $0.12, 0.575, 0.3, 0.0065, 0.55, 0.35, 0.5$ and $1.0$, respectively.

	\section{Experiments}
	We evaluate the proposed method on UVOS DAVIS challenge dataset~\cite{DAVIS_2019}, which contains 30 videos without a mask from the first frame in each video. We directly submit our results to the CodaLab site~\cite{CodaLab} to get segmentation results. Evaluation metrics are Region Jaccard ($\mathcal{J}$) and Boundary $F$ measure ($\mathcal{F}$) for each instance. As shown in~\tabref{tab:results_challenge}, our method achieves the third rank with respect to Global Mean as well as all the other specific metrics.
	
\begin{figure*}
		\begin{center}
			\def\arraystretch{0.5}
			\begin{tabular}{@{}c@{\hskip 0.01\linewidth}c@{\hskip 0.01\linewidth}c@{\hskip 0.01\linewidth}c@{\hskip 0.01\linewidth}c@{\hskip 0.01\linewidth}c@{\hskip 0.01\linewidth}c}
				
				\includegraphics[width=0.192\linewidth]{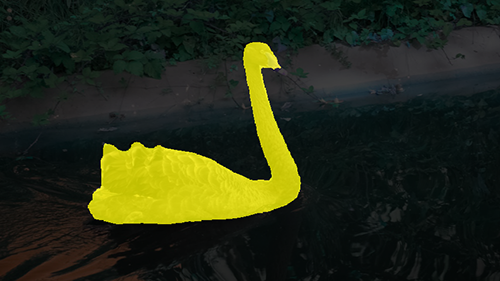} &
				\includegraphics[width=0.192\linewidth]{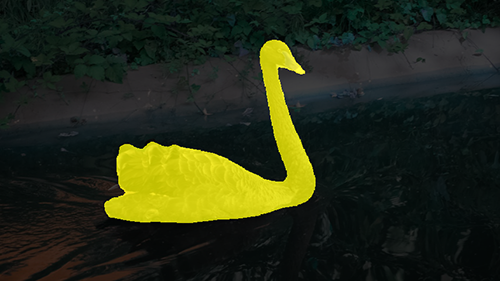} &
				\includegraphics[width=0.192\linewidth]{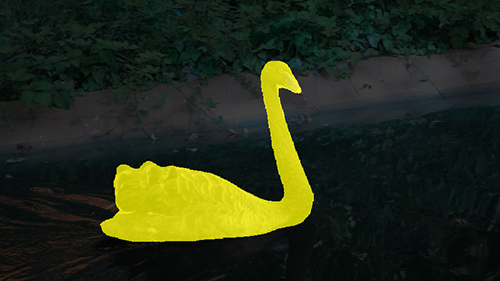} &
				\includegraphics[width=0.192\linewidth]{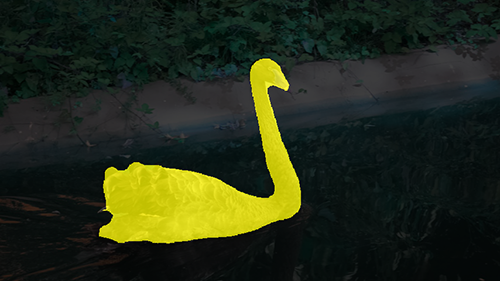} &
				\includegraphics[width=0.192\linewidth]{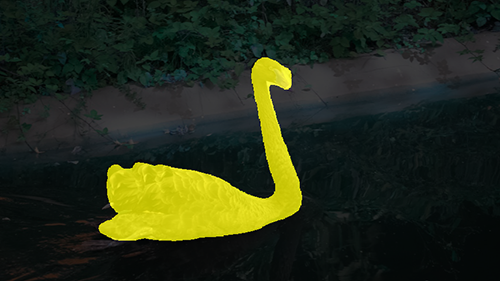}  \\
				\includegraphics[width=0.192\linewidth]{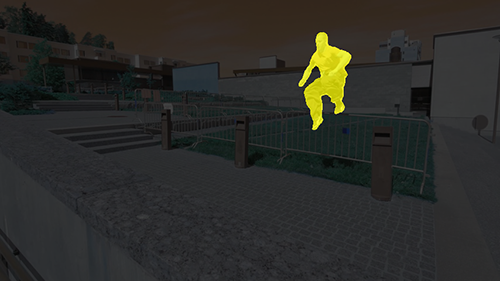} &
				\includegraphics[width=0.192\linewidth]{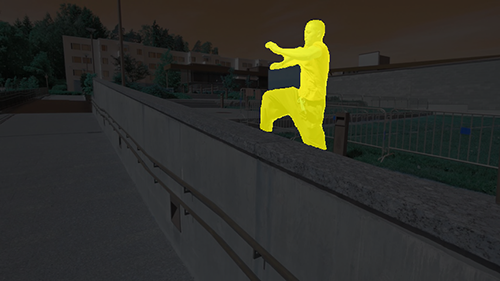} &
				\includegraphics[width=0.192\linewidth]{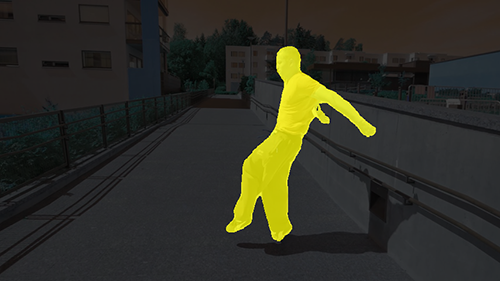} &
				\includegraphics[width=0.192\linewidth]{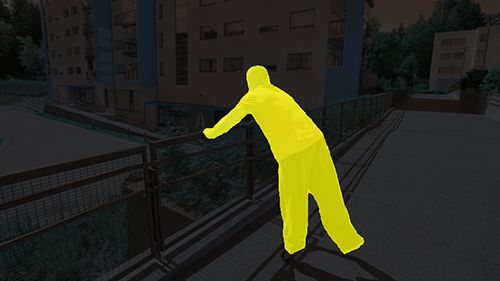} &
				\includegraphics[width=0.192\linewidth]{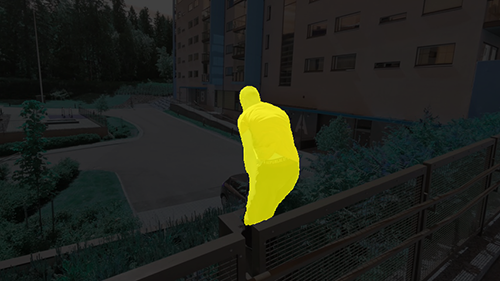}  \\
				\includegraphics[width=0.192\linewidth]{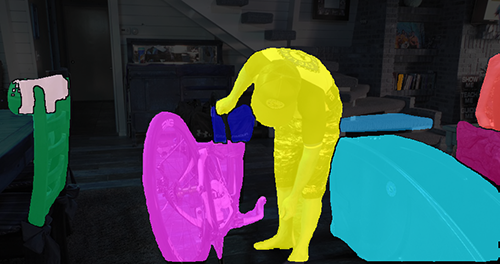} &
				\includegraphics[width=0.192\linewidth]{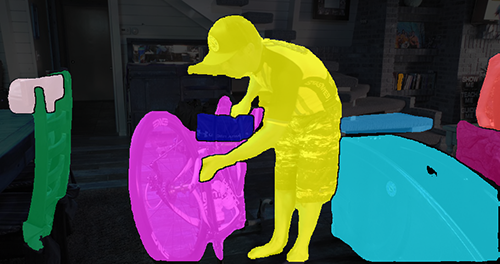} &
				\includegraphics[width=0.192\linewidth]{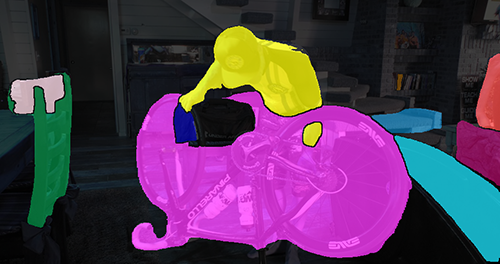} &
				\includegraphics[width=0.192\linewidth]{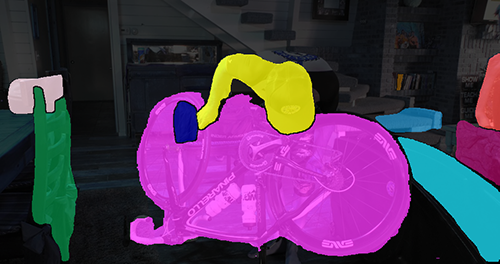} &
				\includegraphics[width=0.192\linewidth]{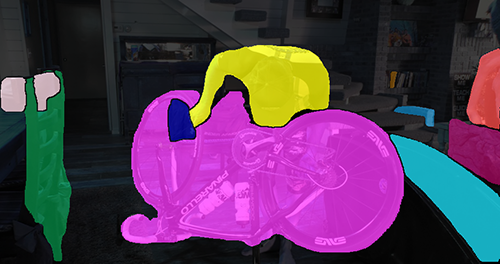}  \\
				\includegraphics[width=0.192\linewidth]{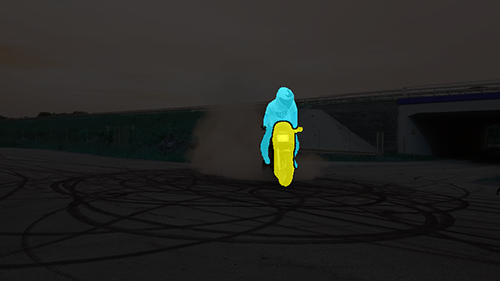} &
				\includegraphics[width=0.192\linewidth]{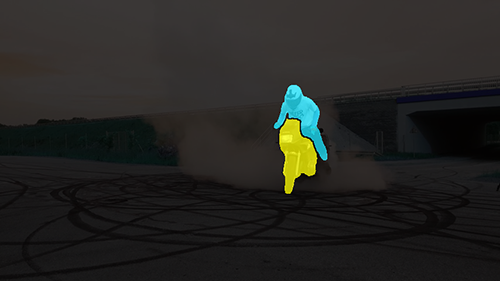} &
				\includegraphics[width=0.192\linewidth]{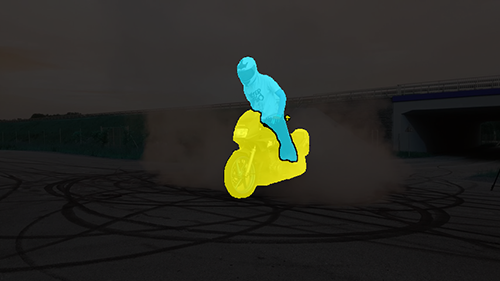} &
				\includegraphics[width=0.192\linewidth]{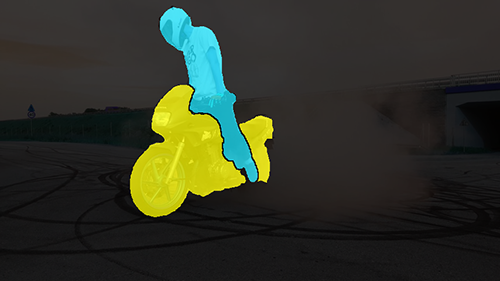} &
				\includegraphics[width=0.192\linewidth]{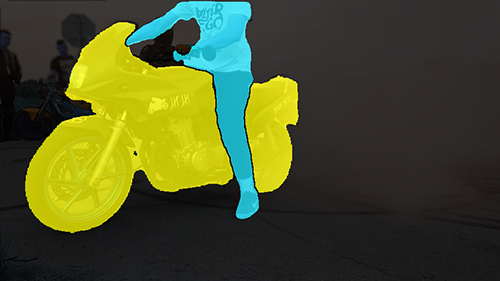}  \\
				\includegraphics[width=0.192\linewidth]{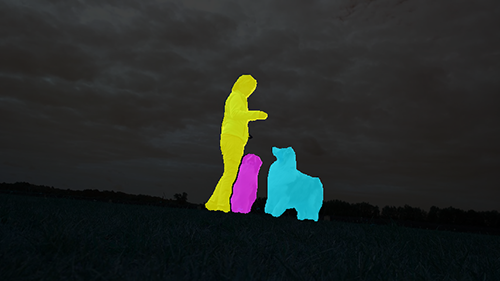} &
				\includegraphics[width=0.192\linewidth]{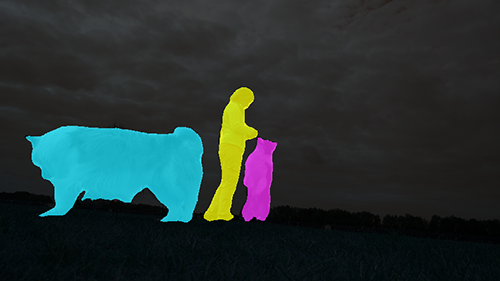} &
				\includegraphics[width=0.192\linewidth]{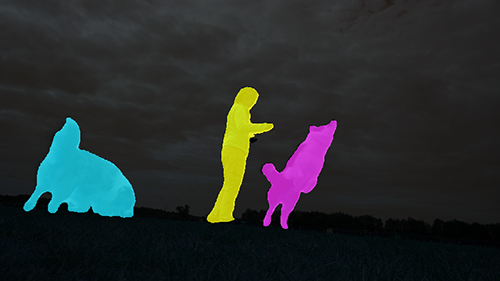} &
				\includegraphics[width=0.192\linewidth]{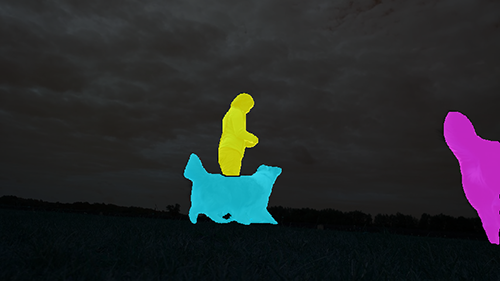} &
				\includegraphics[width=0.192\linewidth]{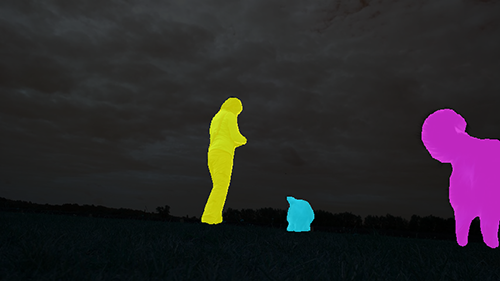}  \\
				\includegraphics[width=0.192\linewidth]{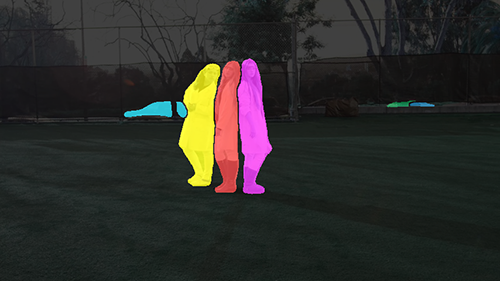} &
				\includegraphics[width=0.192\linewidth]{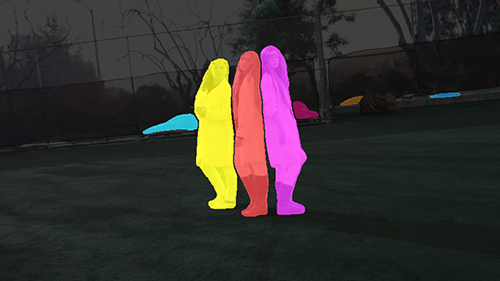} &
				\includegraphics[width=0.192\linewidth]{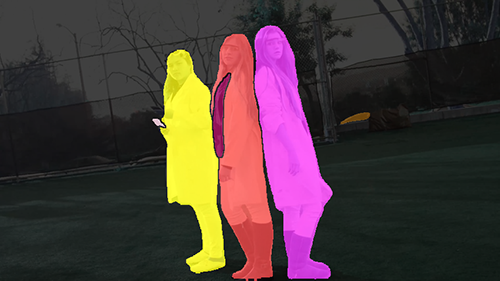} &
				\includegraphics[width=0.192\linewidth]{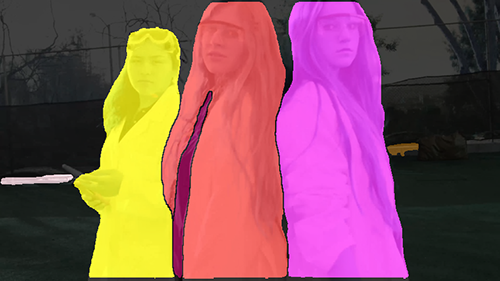} &
				\includegraphics[width=0.192\linewidth]{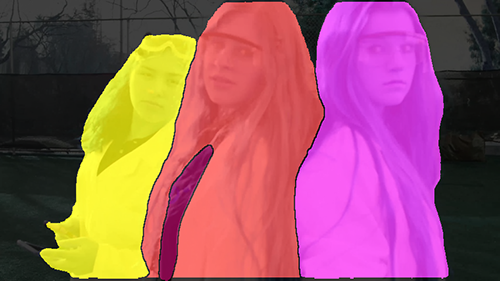}  \\
			\end{tabular}
		\end{center}
		\caption{Qualitative results on DAVIS validation set that  
			contains a single object and multiple objects.}
		\label{fig:Result}
	\end{figure*}

	\figref{fig:Result} shows qualitative results on DAVIS validation set. In the first row of~\figref{fig:Result}, our method well captures~\textit{blackswan} over time. The proposed method also shows robust video segmentation results even in the relatively dynamic~\textit{Parkour} example. In addition, our method faithfully works on multiple object segmentation (from the third row to the last low in~\figref{fig:Result}). Note that changes in scale (\textit{lab-coat}) and appearance (\textit{mbike-trick}) are handled appropriately by our method.
	
	The proposed key instance selection method is quantitatively compared with  random instance selection according to the $K$ as shown in~\tabref{tab:top_k}. 
	Since the proposed scheme focuses on semantically meaningful regions, it shows promising results even at a small $K$.


	\begin{table}[t]
		\centering
		\resizebox{0.9\linewidth}{!}{%
			\begin{tabular}{  l | cc cc cc cc cc cc cc cc}
				\hline
				\# of IDs  && 5 && 10 && 15 && 20 \\
				\hline
				\hline
				Random     && 0.474  && 0.529 && 0.587  && 0.594 \\
				Ours  && 0.576  && 0.579 && 0.591 && 0.599\\
				\hline  
			\end{tabular} 
		}
		\linebreak
		\caption{Global Mean ($\mathcal{J}$$\&$$\mathcal{F}$) scores on the DAVIS validation dataset according to the number of IDs. We compare our key instance selection method with random  instance selection.}
		\label{tab:top_k}
	\end{table}
	
	\section{Conclusion}
	We have presented key instance selection  for  unsupervised video object segmentation (UVOS). Our method allows maximum $K$ instances to be tracked by considering  video saliency and frequency of appearance. Focusing on objects that are in the spotlight enables to reduce time complexity.
	In addition,  objects in a frame sequence are linked by specific scores from  ReID descriptor,  trajectories, and co-segmentation result.
	Finally, we perform hyperparameter search to find out the optimal hyperparameters in our model by Meta AI system.
	Our method showed competitive results in UVOS DAVIS challenge.
	
	\section{Acknowledgements}
	We thank Youngsoon Lee for help to use a Meta AI system which is effective in searching hyperparameters.
	
	{\small
		\bibliographystyle{ieee_fullname}
		\bibliography{egbib}
	}
	
\end{document}